\documentclass[conference]{IEEEtran}

\IEEEoverridecommandlockouts
\usepackage{cite}
\usepackage{amsmath,amssymb,amsfonts}
\usepackage{algorithmic}
\usepackage{graphicx}
\usepackage{textcomp}
\usepackage{xcolor}
\usepackage{booktabs}
\usepackage{multirow}
\usepackage{url}

\begin{document}

\title{PermitGPT: A Unified Generative-AI Pipeline for Construction Hazard Forecasting, Permit Prediction, and Community Impact}

\author{
\IEEEauthorblockN{
Mohd Ruhul Ameen$^{1}$,
Farjana Aktar$^{1}$,
Akif Islam$^{1}$,\\
Momen Khandoker Ope$^{1}$,
Abu Saleh Musa Miah$^{2}$,
Jungpil Shin$^{2}$
}

\IEEEauthorblockA{
ameensunny242@gmail.com,
farjana.aktar.cseru@gmail.com,
iamakifislam@gmail.com,\\
khandokermomen919@ru.ac.bd,
musa@u-aizu.ac.jp,
jpshin@u-aizu.ac.jp
}

\IEEEauthorblockA{
$^{1}$Department of Computer Science and Engineering, University of Rajshahi, Rajshahi 6205, Bangladesh\\
$^{2}$University of Aizu, Aizuwakamatsu, Fukushima, Japan
}
}

\maketitle
\begin{abstract}
Urban construction governance requires early decisions that connect workplace safety, permitting requirements, and community impact, yet the relevant evidence is often scattered across separate municipal and regulatory data sources. This paper presents PermitGPT, a unified generative artificial intelligence framework for converting unstructured construction permit descriptions into structured decision-support outputs across three domains: safety hazard identification, permit requirement specification, and community impact assessment. To address data fragmentation, we spatially and temporally align records from the New York City Department of Buildings, Occupational Safety and Health Administration, and NYC 311 service requests, producing 90,000 structured prompt-response pairs derived through rule-based alignment and domain-informed spot checking. We fine-tune three open-weight language models using parameter-efficient adaptation and evaluate them on 2,833 held-out test cases. The results show complementary model behavior: Gemma-3-1B provides the most efficient inference at 3.07 samples per second with low memory usage, Llama-3.2-3B gives the highest lexical overlap for regulatory-style outputs with a BLEU score of 0.0091, and 4-bit Mistral-7B-Instruct-v0.3 achieves the strongest semantic alignment with a BERTScore-F1 of 0.7747. Because the task involves open-ended structured generation, low BLEU values are interpreted alongside semantic metrics and qualitative output structure rather than as standalone indicators of utility. Overall, PermitGPT provides an initial step toward AI-assisted construction governance while identifying directions for stronger task-level evaluation and real-world validation.
\end{abstract}


\section{Introduction}
Urban construction projects shape the growth of modern cities, but every new project also brings a set of practical risks that must be understood early. A single permit description may imply possible worker hazards, regulatory requirements, temporary disruption to pedestrians, noise complaints, traffic obstruction, or broader community concerns. In practice, however, these signals are rarely examined together. Safety records, permit histories, inspection reports, and citizen complaints are usually stored in separate systems, making it difficult for municipal agencies and project stakeholders to form a complete picture before work begins.

This fragmentation is especially important in dense urban environments such as New York City, where construction activity directly affects workers, residents, businesses, and public infrastructure. Recent statistics from the Occupational Safety and Health Administration show that construction remains a high-risk sector for work-related injuries, while the New York City Department of Buildings reports that permitting and compliance issues contribute to project delays~\cite{OSHA2023Summary,NYCDOB2022Delay}, while the New York City Department of Buildings reports that thirty percent of construction projects encounter delays attributed to permitting and compliance issues~\cite{OSHA2023Summary,NYCDOB2022Delay}. These safety and compliance challenges are not isolated problems. A project with complex work conditions may also require additional permits, create neighborhood disruption, and increase the need for early intervention. Therefore, construction governance requires tools that can connect safety, regulation, and community impact rather than treating them as independent tasks.

Recent advances in large language models (LLMs) and generative artificial intelligence offer a promising direction for this problem. Unlike traditional rule-based systems, LLMs can interpret unstructured project descriptions and generate structured outputs that are useful for decision support. Prior studies have shown that fine-tuned generative models can predict construction accident categories, while AI systems combining computer vision and natural language processing have supported hazard detection and compliance checking~\cite{yoo2024gptSafety,hussain2025isafeinspect}. However, many existing approaches focus on a single objective, such as accident prediction or visual safety monitoring. They often do not provide an integrated view of how a proposed construction activity may affect safety hazards, permit requirements, and surrounding communities at the same time.

To address this gap, this paper presents PermitGPT, a unified generative AI pipeline for early-stage construction risk and impact assessment. PermitGPT transforms raw construction permit descriptions into structured outputs across three connected domains: safety hazard identification, permit requirement specification, and community impact assessment. The framework first integrates heterogeneous municipal and regulatory data sources through spatial and temporal alignment, combining signals from construction permits, safety records, and public service requests. It then fine-tunes three open-weight language models using parameter-efficient adaptation to examine the trade-off between inference efficiency, lexical precision, and semantic coherence.

The main contribution of this work is not to replace human inspectors or regulatory experts, but to provide an early decision-support layer that organizes fragmented information into a more usable form. This study reports text-generation metrics, computational efficiency, and qualitative output structure to provide an initial evaluation of the proposed framework. Together, these results position PermitGPT as a practical step toward AI-assisted municipal construction governance.

\section{Related Work}

\subsection{Generative AI for Construction Safety and Permitting}
\label{sec:related_a}

Recent research has shown that generative artificial intelligence can support construction management by interpreting unstructured documents, summarizing safety records, and preparing decision-support outputs. Large Language Models (LLMs) are especially useful in this context because many construction documents, including permits and safety reports, are written in natural language rather than in clean structured formats~\cite{wan2024generativeAI}. Prior work has also combined language-based reasoning with computer vision for safer site monitoring, including hazard detection, protective equipment checking, and rapid feedback to safety managers~\cite{usama2024aiSafety,tang2024riskAssessment}.

Yoo \textit{et al.}~\cite{yoo2024gptSafety} showed that a GPT-based model could classify six types of construction accidents with 82\% accuracy, while Hussain \textit{et al.}~\cite{hussain2025isafeinspect} proposed \textit{iSafeInspect}, an intelligent safety inspection framework that combines visual assessment with practical safety guidance. These studies show the value of AI for construction safety. However, most existing systems focus on a specific task, such as accident classification, visual hazard detection, or inspection assistance, rather than jointly considering safety, permitting, and community impact.

\subsection{Efficient Adaptation and Deployment of Language Models}

For construction applications, general-purpose LLMs usually require adaptation because the language of permits, safety incidents, and regulatory documents is domain-specific. Parameter-efficient methods such as Low-Rank Adaptation (LoRA) and quantization reduce the computational cost of model adaptation while preserving useful task performance~\cite{zhao2024lora,hsu2024safe}. Safety-Aware Fine-Tuning (SAFT) and Safe LoRA further emphasize the importance of aligning model outputs with safety requirements and responsible deployment practices~\cite{choi2024safety}. Efficient optimization strategies have also been explored for reducing computational overhead in construction management contexts~\cite{parthasarathy2024efficient}.

These methods are important for practical deployment because municipal agencies and smaller organizations may not have access to large-scale computing infrastructure. Related work on fine-tuning and domain adaptation suggests that task-specific customization can improve the usefulness of LLMs in specialized professional settings~\cite{weng2024navigating,lu2025fine}. In this paper, we follow this direction by using parameter-efficient adaptation to compare three open-weight models under the same PermitGPT pipeline.

\subsection{Evaluation, Ethics, and Responsible Use}

The evaluation of AI systems in construction remains challenging because different tasks require different measures of success. Classification-oriented studies often report accuracy, precision, recall, F1-score, MAE, or RMSE~\cite{jiao2024safety}. Incident analysis studies and benchmark datasets based on OSHA reports or construction records provide useful foundations for evaluating predictive safety models~\cite{awolusi2022incident}. However, generative systems introduce additional complexity because multiple valid textual responses may describe the same permit-related risk. As a result, lexical metrics such as BLEU may not fully capture practical usefulness, and semantic or task-specific evaluation is often needed.

Responsible deployment is also important. AI-based safety and permitting tools must respect construction regulations, data privacy, fairness, accountability, and the limits of automated decision-making~\cite{usama2024aiSafety,raliile2020machine,behzadan2024formalizing,ajirotutu2024ai}. Prior studies emphasize that such systems should support, not replace, human expertise, and should involve collaboration among engineers, regulators, safety experts, and community stakeholders~\cite{ruchit2024progress,huang2024ai}. PermitGPT follows this view by framing its outputs as early decision-support information rather than final regulatory judgments.

\subsection{Research Gap and Position of PermitGPT}

Existing literature has made meaningful progress in construction safety prediction, visual hazard detection, AI-assisted inspection, and efficient model adaptation. Researchers have also highlighted the need to combine multiple sources of construction data for stronger risk assessment and warning systems~\cite{gao2024fusing,kamil2024multi,kumi2024systematic}. Graph-based models, transfer learning, and hybrid human--AI systems offer further opportunities for predictive analytics and real-time monitoring~\cite{junjia2024transferLearning,mostofi2023construction,sargiotis2024transforming}.

Despite these advances, three gaps remain important. First, many studies focus on a single task, while construction governance often requires simultaneous consideration of safety hazards, permit requirements, and community impact. Second, early-stage assessment remains difficult because relevant evidence is fragmented across permit records, safety data, and citizen complaint systems. Third, relatively few works examine lightweight generative models as practical tools for converting permit descriptions into structured multi-domain outputs.

PermitGPT addresses these gaps by integrating heterogeneous municipal and regulatory data sources and adapting open-weight LLMs for structured construction decision support. Rather than claiming to solve all aspects of construction risk management, this work provides an initial framework for connecting fragmented early-stage information and supporting future task-specific evaluation, baseline comparison, and real-world validation.

\section{Methodology}

\begin{figure*}[!t]
    \centering
    \includegraphics[width=0.8\textwidth]{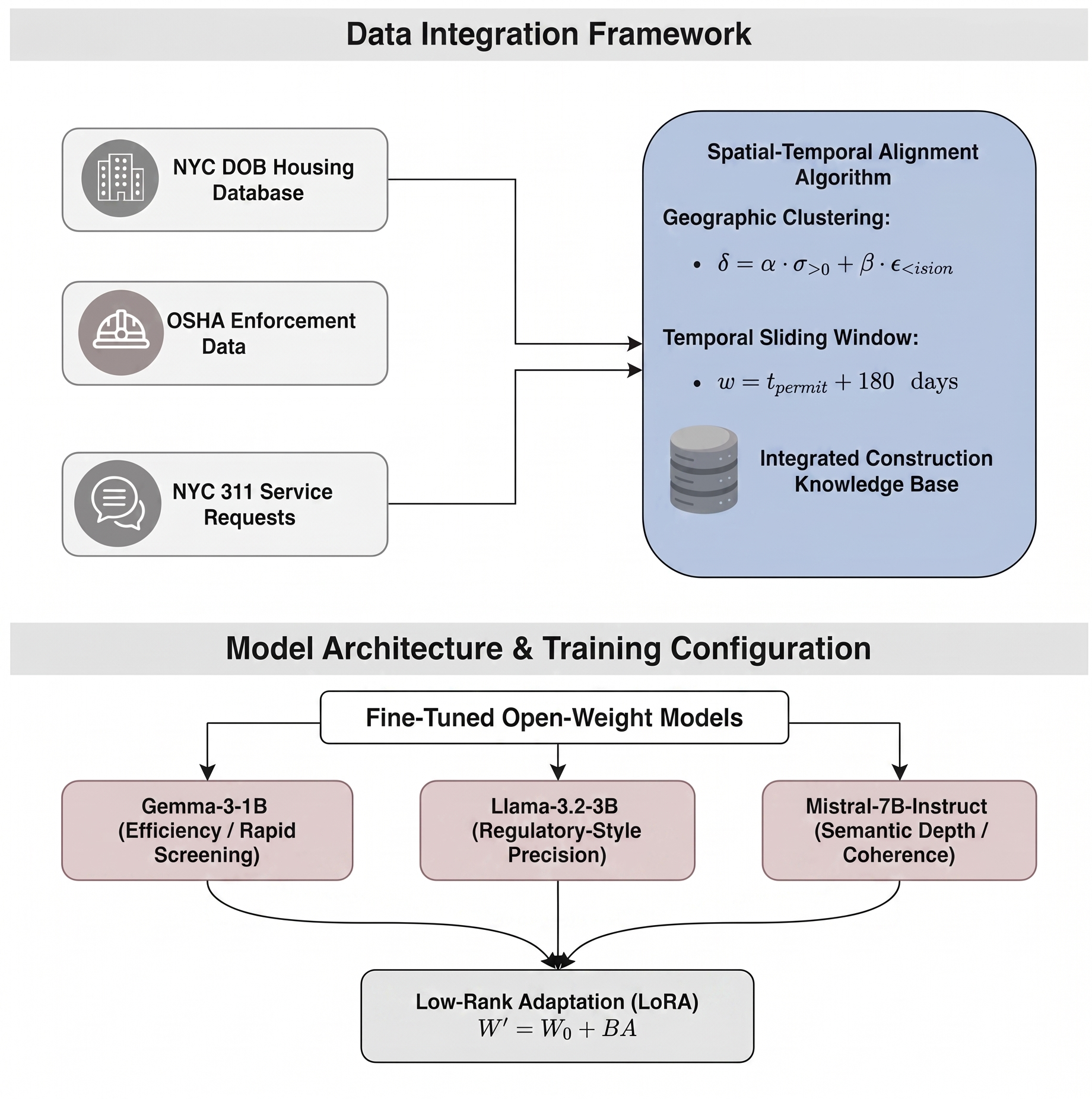}
    \caption{Proposed PermitGPT pipeline. The three model branches correspond to Gemma-3-1B for efficiency-oriented rapid screening, Llama-3.2-3B for regulatory-style lexical precision, and Mistral-7B-Instruct-v0.3 for semantic depth and coherence.}
    \label{fig:permitgpt_pipeline}
\end{figure*}

\subsection{Data Integration Framework}

PermitGPT is designed to connect construction-related information that is normally stored in separate systems. The framework uses three municipal and regulatory data sources: New York City Department of Buildings records, OSHA Enforcement Data, and NYC 311 service requests. Together, these sources provide complementary views of a construction project, including permit descriptions, safety-related incidents, and community-level complaints.

The NYC Department of Buildings Housing Database serves as the primary source of construction permit information. It contains more than 1.2 million permit records dating from January 2010~\cite{NYCDCP_Housing_Database}. These records include structured fields such as project type, location, and contractor information, as well as free-text job descriptions that often contain important project-specific details. OSHA Enforcement Data provides the safety-related context through workplace inspection and violation records~\cite{USDOL_OSHA_EnforceData}. Construction-related cases were identified using North American Industry Classification System codes 236--238. NYC 311 service requests were used to capture community-facing signals such as noise, sidewalk obstruction, dust, and access disruption.

The data integration process followed a rule-based spatial and temporal alignment strategy. Permit locations were matched with nearby OSHA incidents and 311 complaints using geographic proximity, while temporal matching associated events with the active permit period and an additional 180-day window to capture delayed safety or community impacts. This process produced structured prompt-response pairs in which each permit description was connected to three output categories: likely safety hazards, possible permit requirements, and anticipated community impacts. The final corpus was constructed through rule-based alignment and domain-informed spot checking rather than full manual double annotation.

\subsection{Model Selection and Fine-Tuning}

Three open-weight language models were selected to represent different deployment trade-offs. Gemma-3-1B was included as a lightweight model for rapid screening and lower memory usage. Llama-3.2-3B was selected as a mid-sized model expected to provide stronger lexical alignment for regulatory-style text. Mistral-7B-Instruct-v0.3 was included as a larger instruction-tuned model for stronger semantic coherence and more detailed responses.

All models were fine-tuned using Low-Rank Adaptation (LoRA)~\cite{Hu2022LoRA}, which updates a small number of trainable adapter parameters while keeping the original model weights frozen. The LoRA update is defined as:
\begin{equation}
W' = W_0 + BA
\end{equation}
where $W_0$ represents the frozen pretrained weights, and $B$ and $A$ are trainable low-rank matrices. In this study, the LoRA rank was set to $r=8$, allowing the models to adapt to the construction-domain task with limited computational overhead.

Training used the AdamW optimizer with a learning rate of $2 \times 10^{-4}$, weight decay of 0.01, warmup, and a linear decay schedule. This configuration was chosen to support efficient parameter adaptation while keeping the fine-tuning process lightweight. The same general training setup was applied across the three models to make their performance and efficiency more comparable.

\subsection{Output Structure and Evaluation Setup}

Each model was trained to convert an input permit description into a structured response with three components: safety hazard identification, permit requirement specification, and community impact assessment. This design reflects the practical goal of PermitGPT: to organize early construction information into a form that can support human review.

The models were evaluated on 2,833 held-out test cases. Because the task is generative rather than a fixed-label classification problem, evaluation used text-generation metrics that capture both lexical and semantic similarity. BLEU, ROUGE-1, and METEOR were used to measure surface-level overlap with reference responses, while BERTScore was used to measure semantic alignment. In addition, inference efficiency was considered to understand the practical deployment trade-off among the three models.

\section{Results and Discussion}

\subsection{Quantitative Performance Evaluation}

Table~\ref{tab:performance} summarizes the performance of the three fine-tuned models across the training, validation, and test splits. Since PermitGPT generates structured text rather than fixed class labels, several evaluation metrics are reported to capture different aspects of output quality. BLEU and ROUGE-1 measure lexical overlap, METEOR considers partial word matching and synonym-level similarity, and BERTScore measures semantic similarity using contextual embeddings.

\begin{table}[!t]
\centering
\caption{Comparative Performance Across Data Splits}
\label{tab:performance}
\begin{tabular}{@{}llcccc@{}}
\toprule
Model & Split & BLEU & BERTScore & ROUGE-1 & METEOR \\
\midrule
\multirow{3}{*}{Gemma-3-1B} 
& Train & 0.0025 & 0.7709 & 0.1166 & 0.0206 \\
& Valid & 0.0026 & 0.7709 & 0.1164 & 0.0206 \\
& Test & 0.0024 & 0.7703 & 0.1154 & 0.0206 \\
\midrule
\multirow{3}{*}{Llama-3.2-3B} 
& Train & 0.0093 & 0.7747 & 0.1486 & 0.0376 \\
& Valid & 0.0095 & 0.7749 & 0.1490 & 0.0378 \\
& Test & \textbf{0.0091} & 0.7742 & \textbf{0.1470} & \textbf{0.0379} \\
\midrule
\multirow{3}{*}{Mistral-7B} 
& Train & 0.0024 & 0.7752 & 0.1169 & 0.0205 \\
& Valid & 0.0025 & 0.7753 & 0.1168 & 0.0205 \\
& Test & 0.0024 & \textbf{0.7747} & 0.1156 & 0.0205 \\
\bottomrule
\end{tabular}
\end{table}

The results show that the models behave differently across lexical and semantic metrics. Llama-3.2-3B obtains the highest test BLEU, ROUGE-1, and METEOR scores, suggesting that it produces outputs with relatively stronger surface-level overlap with the reference responses. Mistral-7B-Instruct-v0.3 achieves the highest test BERTScore, indicating stronger semantic alignment even when the generated wording differs from the reference. Gemma-3-1B performs slightly lower on text-quality metrics but remains useful because of its computational efficiency, as discussed later.

The BLEU scores are low because the task involves open-ended structured generation rather than exact sentence reproduction. Multiple valid responses may describe the same hazard, permit requirement, or community impact using different wording. For context, BLEU scores below 0.05 are common in open-domain generation tasks such as abstractive summarization and dialogue systems, where diverse valid phrasings exist for the same underlying meaning~\cite{reiter2018bleu}. Therefore, lexical overlap metrics such as BLEU and METEOR are less informative for this task than semantic similarity metrics. BERTScore computes similarity using contextual embeddings and is more tolerant of valid paraphrase variation; it is therefore treated as the main indicator of semantic alignment in this evaluation.

The stability of the metrics across training, validation, and test splits also suggests that the models did not show large split-level performance fluctuations. The maximum absolute deviation between training and test scores remains small for most metrics, which indicates reasonably consistent behavior across the evaluated data splits. Although this study does not include a separate zero-shot baseline, the comparison among three fine-tuned open-weight models provides an initial assessment of efficiency, lexical overlap, and semantic coherence under the proposed pipeline.

\subsection{Qualitative Output Analysis}

To make the generated output easier to interpret, Table~\ref{tab:qualitative_example} presents a representative example of the PermitGPT response structure. The example is included to illustrate the practical organization of the generated outputs rather than to serve as an additional quantitative benchmark.

\begin{table}[!t]
\centering
\caption{Representative PermitGPT Output Structure}
\label{tab:qualitative_example}
\begin{tabular}{p{0.95\linewidth}}
\toprule
\textbf{Input permit description:} Interior renovation involving demolition, electrical rewiring, and temporary sidewalk obstruction.\\
\midrule
\textbf{Generated safety hazards:} Possible demolition debris exposure, electrical hazards, trip hazards, and worker/public separation risks.\\
\textbf{Generated permit requirements:} Electrical permit review, demolition safety compliance, sidewalk obstruction approval, and site protection measures.\\
\textbf{Generated community impact:} Temporary pedestrian disruption, noise, dust, and possible access limitations near the worksite.\\
\bottomrule
\end{tabular}
\end{table}

This example shows the intended role of PermitGPT as an early decision-support tool. Instead of producing a single label, the system organizes a short permit description into three practical categories: safety, permitting, and community impact. Such a structure can help reviewers quickly identify issues that may require closer human inspection. At the same time, the example also shows why exact lexical matching is difficult for this task. The same meaning can be expressed using different valid phrases, which reinforces the need to interpret BLEU and other overlap-based metrics with caution.

\subsection{Computational Efficiency and Deployment Considerations}

In addition to output quality, computational efficiency is important for municipal and construction-management settings where resources may be limited. Among the evaluated models, Gemma-3-1B achieved the highest inference efficiency at approximately 3.07 samples per second with comparatively low VRAM usage. This makes lightweight models attractive for rapid screening or preliminary review scenarios. In contrast, Mistral-7B-Instruct-v0.3 produced the strongest semantic alignment but required higher computational resources, reflecting the usual trade-off between model capacity and inference cost.

The complementary behavior of the three models suggests that PermitGPT can support different operational needs. A smaller model may be useful when many permit descriptions need to be processed quickly, while a larger model may be preferred when semantic detail and interpretability are more important. Llama-3.2-3B occupies a middle position by offering stronger lexical overlap for regulatory-style responses. This tiered interpretation is useful because practical AI deployment rarely depends on a single metric alone; it depends on the balance among speed, memory use, consistency, and output quality.

Training behavior was also stable during fine-tuning. The observed loss trends decreased smoothly without major instability, suggesting that the LoRA-based adaptation strategy was adequate for the present domain adaptation setting. Overall, the results indicate that parameter-efficient fine-tuning can provide a practical balance between construction-specific language understanding and computational feasibility.

\subsection{Human-in-the-Loop Use and Quality Assurance}

PermitGPT is intended to support, not replace, human decision-makers. Construction safety and permitting decisions involve legal, technical, and community-sensitive considerations, and therefore generated outputs should be treated as decision-support information rather than final regulatory judgments. In a practical workflow, the system may first organize a permit description into likely hazards, permit needs, and community impacts, after which inspectors or municipal staff can review and refine the output.

A human-in-the-loop design is especially important because generative models may sometimes produce incomplete, overly general, or uncertain responses. For sensitive cases, confidence-aware routing and manual verification can help improve accountability. Lightweight models may be used for rapid first-pass screening, while more detailed model outputs or human review may be reserved for complex projects. This type of workflow keeps the system practical while preserving the role of expert judgment.
\subsection{Limitations and Future Directions}

\begin{table}[!b]
\centering
\caption{Evaluation Scope and Current Limitations}
\label{tab:evaluation_scope}
\begin{tabular}{p{0.38\linewidth}p{0.52\linewidth}}
\toprule
\textbf{Aspect} & \textbf{Status in This Study} \\
\midrule
Model comparison & Three fine-tuned open-weight models are compared. \\
Zero-shot baseline & Not included in the current version. \\
Task-specific accuracy & Not separately computed for hazard, permit, and community-impact outputs. \\
Ablation study & Not conducted for individual data sources or pipeline components. \\
Qualitative output & A representative structured output example is provided. \\
Human review & Recommended for sensitive permitting and safety decisions. \\
\bottomrule
\end{tabular}
\end{table}

Table~\ref{tab:evaluation_scope} summarizes the main evaluation boundaries of this study. The current work compares three fine-tuned open-weight models, but it does not include zero-shot baselines, task-specific accuracy for each output category, or component-level ablation. Therefore, the performance gains attributable specifically to fine-tuning and to individual data sources cannot be quantified from the present results alone. In addition, the corpus was produced through rule-based alignment with domain-informed spot checking rather than full double annotation, so formal inter-annotator agreement was not measured.

These limitations should be considered when interpreting the reported results. Future work will address them through controlled baseline comparisons, task-level evaluation with structured output parsing, component-wise ablation studies, and prospective real-world case studies. Future versions of PermitGPT should also explore multimodal inputs, such as construction plans and site images, along with real-time sensor data and federated learning for cross-jurisdictional deployment. These directions would help evaluate whether the proposed framework can move from retrospective analysis toward practical use in live construction governance workflows.

\section{Conclusion}
This paper presented PermitGPT, a unified generative AI framework for organizing fragmented construction-related information into structured decision-support outputs. By integrating permit records, safety data, and community service requests, the framework connects three practical concerns in urban construction governance: safety hazard identification, permit requirement specification, and community impact assessment. The evaluation of three fine-tuned open-weight language models shows complementary behavior across efficiency, lexical overlap, and semantic alignment, suggesting that different models may serve different operational needs.

Rather than replacing human inspectors or regulatory experts, PermitGPT is intended as an early-stage assistive tool that can help organize information for further review. The current study provides an initial evaluation of this direction, while also leaving important work for future research, including zero-shot baseline comparison, task-specific accuracy measurement, component-wise ablation, and real-world case validation.

\bibliographystyle{IEEEtran}
\bibliography{ref.bib}

\end{document}